\documentclass[runningheads, envcountsame, a4paper]{llncs}
\usepackage{xcolor}
\usepackage{times}
\usepackage{epsfig}
\usepackage{graphicx}
\usepackage{amsmath}
\usepackage{amssymb}
\usepackage{svg}
\usepackage{amsmath}
\usepackage{booktabs}
\usepackage{bm}
\usepackage{upgreek}
\usepackage[misc]{ifsym}
\usepackage{epstopdf}
\usepackage[hyphens]{url}
\makeatletter
\usepackage{microtype}
\DeclareRobustCommand\onedot{\futurelet\@let@token\@onedot}
\def\@onedot{\ifx\@let@token.\else.\null\fi\xspace}

\begin{document}

\title{Pairwise-GAN: Pose-based View Synthesis through Pair-Wise Training}

\author{Xuyang Shen \and Jo Plested 
\and Yue Yao \and Tom Gedeon}
\authorrunning{X. Shen et al}
\institute{Research School of Computer Science,\\ Australian National University \\
\email{first.second@anu.edu.au}}
\maketitle              % typeset the header of the contribution
\begin{abstract}
Three-dimensional face reconstruction is one of the popular applications in computer vision. However, even state-of-the-art models still require frontal face as inputs, which restricts its usage scenarios in the wild. A similar dilemma also happens in face recognition. New research designed to recover the frontal face from a single side-pose facial image has emerged. The state-of-the-art in this area is the Face-Transformation generative adversarial network, which is based on the CycleGAN. This inspired our research which explores the performance of two models from pixel transformation in frontal facial synthesis, Pix2Pix and CycleGAN. We conducted the experiments on five different loss functions on Pix2Pix to improve its performance, then followed by proposing a new network Pairwise-GAN in frontal facial synthesis. Pairwise-GAN uses two parallel U-Nets as the generator and PatchGAN as the discriminator. The detailed hyper-parameters are also discussed. Based on the quantitative measurement by face similarity comparison, our results showed that Pix2Pix with L1 loss, gradient difference loss, and identity loss results in 2.72$\%$ of improvement at average similarity compared to the default Pix2Pix model. Additionally, the performance of Pairwise-GAN is 5.4$\%$ better than the CycleGAN and 9.1$\%$ than the Pix2Pix at average similarity.

\end{abstract}
\section{Introduction}
\label{cha:intro}

Three-dimensional face reconstruction from a single image is one of the popular topics in computer vision over the past twenty years. Although end-to-end learning methods achieve the state-of-the-art in three-dimensional facial reconstruction~\cite{jackson2017large,feng2018joint}, they still cannot infer the missing facial information if the input facial image is a side-pose at more than 30 degrees. This issue not only happens on three-dimensional facial reconstruction tasks but also exists on other facial applications in computer vision.  Face recognition is a widely used authentication and detection technique that currently requires front face images. However, this condition is challenging to fulfill in real life, especially for video surveillance which captures the object under any situation.

Inferring the missing facial information, known as frontal facial synthesis, is a good alternative for these applications to solve the issue raised by the side-pose facial image. From 2013, an increasing number of deep learning methods have been proposed to improve performance on frontal face synthesis problems~\cite{zhang2013random,kan2014stacked,hassner2015effective,yin2017towards}. With the publication of Two Pathways Generative Adversarial Network (TP-GAN)~\cite{huang2017beyond} and complete representations generative adversarial network (CR-GAN)~\cite{tian2018cr}, the focus of solving face synthesis problems was shifted into models based on GAN and CGAN~\cite{huang2017beyond,tian2018cr,zhuang2019ft,yin2020dual}. 

The current state-of-the-art frontal facial synthesis is Pose-Invariant Generator Adversarial Network (FT-GAN) proposed by~\cite{zhuang2019ft}. FT-GAN combines Cycle-Consistent Adversarial Networks (CycleGAN), considered state-of-the-art in pixel-to-pixel transformation~\cite{zhu2017unpaired,yao2019simulating,yao2020information,tewari2020state}, with key point alignment to generate frontal facial synthesis more realistically. Their results indicated that FT-GAN is 10$\%$ and 4$\%$ better than CR-GAN and CycleGAN respectively, when the input side-pose image is between 60 and 75 degrees.

With these motivations, we begin our work by exploring two models, Pix2Pix and CycleGAN, from style transformation into frontal facial synthesis. Our research aimed to evaluate the performance of these two models under default hyper-parameters using the Color FERET Database~\cite{nistgov}. We conducted our experiments in two stages:
\begin{enumerate}
\item We applied different loss functions to the Pix2Pix model and evaluated their performance.
\item We proposed a new CGAN network named Pairwise-GAN, based on the network architecture of Pix2Pix and CycleGAN, and compared its performance with state-of-the-art models. 
\end{enumerate}

The main contributions of this work are listed below:
\begin{itemize}
\item Five different loss functions were analyzed to improve the performance of the Pix2Pix in frontal facial synthesis. The best loss results achieved an 2.72$\%$ improvement compared to Pix2Pix, which is close to CycleGAN's performance. 
\item Pairwise-GAN which targets frontal facial synthesis is proposed as a new network architecture of CGAN. Pairwise-GAN reaches 44.3 average similarity and 74.22 maximum similarity between synthesis and ground truth. It gains 5.4$\%$ better results than the CycleGAN and 9.1$\%$ than the Pix2Pix. Compared to a 4$\%$ improvement over CycleGAN by the FT-GAN~\cite{zhuang2019ft}.
\item A new quantitative measurement on face frontalization is introduced to evaluate the similarity between the generated frontal face and the ground-truth image. 
\end{itemize}

\section{Methods}
\label{cha:methodology}
\subsection{Dataset and Pre-processing}

We selected the Color FERET Database from NIST~\cite{nistgov} since it is free to access for research purposes. It contains 11,338 facial images from 994 different subjects. Images were collected at 14 different times between December 1993 to August 1996~\cite{nistgov}, so the focus length and position of the camera are different each time. This results in various facial sizes. To minimize the influence of this noise to GAN training, we used MTCNN ~\cite{zhang2016joint} to crop out the face for each image.

Another source of variability of the dataset is the angle of side-images. It is commonly agreed that synthesizing the frontal image based on the side-image at a small angle is more accessible than the large-angle since the former one can provide more facial features. Therefore, we added another constraint to frontal facial generation by requiring the angle of side-pose images to be at least 60 degrees. We also restricted the maximum angle of fewer than 90 degrees as models gain fewer facial information after this degree.

\subsubsection{Train set and Test set}

After pre-processing, there are 3,135 images (2,090 pairs) in the training set and 369 (246 pairs) in the test set, where one pair contains one frontal image and one side-pose image. The size of input images in experiments is further resized into 256 by 256 pixel, due to the limitation of computation. The reason we chose 256 is that elaborate facial features can be reflected in the generated image from 256 to 512 pixels, such as hairstyle or detailed information on eyes~\cite{karras2019style}. 

\subsection{Synthesis Loss Functions}

\subsubsection{Experimental setup}

\begin{figure}[t]
      \centering
      \includegraphics[height=2cm]{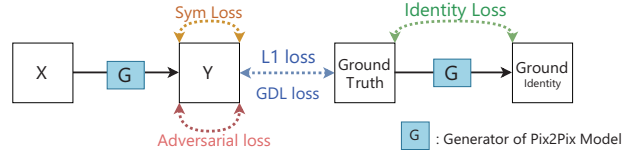}
      \caption{Overview of loss functions used in "exploring loss function" experiment. $X$ and $Y$ in the figure are the input image (side-pose facial image) and the generated frontal image. The ground identity is generated by applying the generator to the ground truth image}
      \label{fig:loss_pix2pix}
\end{figure}

Based on the train set and test set, we performed our first stage of experiments, which explored five different loss functions on Pix2Pix~\cite{isola2017image}. The overview of four different loss functions is shown in Fig.~\ref{fig:loss_pix2pix}. This model was chosen because Pix2Pix and CycleGAN are two state-of-art models in style transformation where frontal facial synthesis also belongs. Additionally, the Pix2Pix is more applicable to experiment with different loss functions compared to the CycleGAN.

The generator in Pix2Pix is a modified U-Net~\cite{ronneberger2015u,isola2017image} which contains eight blocks in the encoder and eight blocks in the decoder. Each block in the encoder is constructed by one convolution layer, one batch normalization layer, and Leaky ReLU as the activation function. Each block in the decoder consists of one transposed convolution layer, one batch normalization layer, and ReLU as the activation function. Additionally, the dropout is also applied to the first three blocks of the decoder.

 \subsubsection{Adversarial loss}
aims to optimize the generator, which achieves a minimum Kullback-Leibler divergence (KL-divergence) between the generated data and ground truth data~\cite{goodfellow2014generative}. Theoretically, the generator tries to get a higher score from the discriminator. In other words, the generated image trends to be a sharp frontal facial image whose facial feature is closer to the input person. However, in practice, it commonly appears that the discriminator in the conditional adversarial network only focuses on whether the generated image is sharp and contains basic facial features or not, ignoring the input of the domain. Based on this, we continued to explore other loss functions. 

\begin{equation}
  Adversarial \ Loss = E_{x\sim P_{x}(X)}[log( D(G(x)))]
\end{equation}  
where $G(x)$ refers to generated data (fake image) by the generator; $D$ refers to the discriminator in Pix2Pix; $G$ refers to the generator in Pix2Pix mode.

\subsubsection{L1 Loss (Mean Absolute Error)}

is employed to facilitate content consistency between the generated frontal image and the ground-truth frontal image. L1 loss also accelerates optimization in CGAN~\cite{huang2017beyond}. 

\begin{equation}
    L1\ Loss = \frac{1}{W \times H} \sum_{i=1}^{W} \sum_{j=1}^{H} |\ GT_{i,\ j}  - Y_{i,\ j}\ | 
\end{equation}
where $GT$ refers to the ground truth image;  $Y$ refers to the generated image; $W$, $H$ is the width and height of the images; $i$, $j$ is the $x$ coordinate and $y$ coordinate of each pixel.

 \subsubsection{Gradient Difference Loss (GDL)}

aims to penalize the differences of image gradient predictions directly, where gradients are the differences between neighboring pixel values~\cite{mathieu2015deep}. Since the facial images usually have continuous value within the neighboring, this formula can strengthen this relationship in the generated image, making it more realistic.

\begin{align}
\begin{split}
    GDL\ Loss = \sum_{i,\ j} ({}& ||Y_{i,\ j} - Y_{i-1,\  j}|- |GT_{i,\ j} - GT_{i-1,\  j}||^\alpha + \\
    & ||Y_{i,\ j-1} - Y_{i,\  j}|- |GT_{i,\ j-1} - GT_{i,\  j}||^\alpha )
\end{split}
\end{align}
where $Y$, $GT$ represents the generated image and ground truth image respectively; $i$, $j$ is the $x$ coordinate and $y$ coordinate of that pixel; $\alpha$ can be any integer greater or equal to $1$, we tested $\alpha = 1 \ and \ \alpha = 2$ in our experiments.

\subsubsection{Symmetry Loss}

Symmetry is one of the features in human faces. Based on this phenomenon, ~\cite{huang2017beyond} proposed the symmetry loss, which encourages symmetrical structure generated by the generator. Additionally, this formula behaves more robustly in the Laplacian image space as it can overcome the illumination difference between two sides of faces. However, we only apply it to the original pixel image, because the illumination difference is minor in Color FERET dataset. 
 
 \begin{equation}
  Symmetry \ Loss = \frac{1}{W/2 \times H} \sum_{i=1}^{W/2} \sum_{j=1}^{H} |\ Y_{i,\ j} - Y_{W-(i-1),\ j}\ | 
\end{equation}
where $Y$ refers to the generated image; $W$, $H$ is the width and height of the images; $i$, $j$ is the $x$ coordinate and $y$ coordinate of that pixel.

\subsubsection{Identity Loss}

 first appeared in the implementation of CycleGAN~\cite{zhu2017unpaired} and was discussed in Cycada~\cite{hoffman2018cycada}. One intention of this function is to regularize the generator, which should not map the input image into a different domain image if it is already in the target domain. Our experiment found that this loss function helps preserve facial identity information from different people. 

\begin{equation}
  Identity \ Loss = \frac{1}{W \times H} \sum_{i=1}^{W} \sum_{j=1}^{H} |\ GT_{i,\ j}  - GI_{i,\ j}\ | 
\end{equation}
where $GT$ refers the ground truth image;  $GI$ refers to the ground identity image; $W$, $H$ is the width and height of the images; $i$, $j$ is the $x$ coordinate and $y$ coordinate of that pixel.

\subsection{Network Architecture of Pairwise-GAN}

During the experiments, we discovered that Pix2Pix and CycleGAN have outstanding achievements in general pixel-to-pixel transformation but gain ordinary performance on frontal facial synthesis. Notably, it was hard to balance the ratio of adversarial loss and other types of losses. If the proportion of adversarial loss is adjusted higher than the other,  the generator can easily deceive the discriminator by generating a realistic image unrelated to the input domain. Alternatively, parts of the generated image are close to the original person but the entire image is not a real face, for example, the face may contain three eyes. As a result of further study of the network architecture, we developed Pairwise-GAN which is based on the network architecture of Pix2Pix and CycleGAN.  

\subsubsection{Generator}
\begin{figure}[t]
      \centering
      \includegraphics[height=4.2cm]{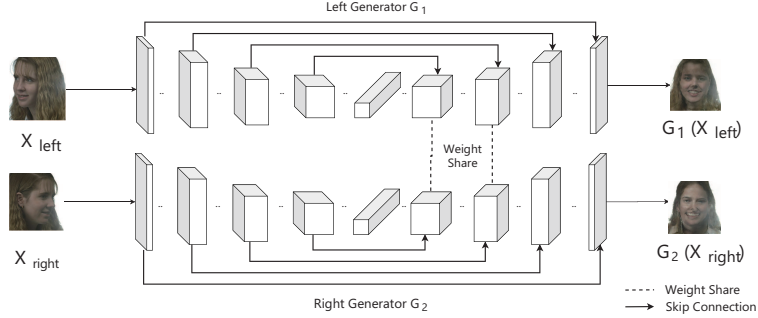}
      \caption{Generator Architecture of Pairwise-GAN. $X_{left}$ left-side facial input; $X_{right}$ right-side facial input. Providing both side faces is a common attribute for most facial databases, such as Multi-PIE Face Database, CAS-PEAL Face Database, and Color FERET Database}
      \label{fig:pw_gan}
\end{figure}

The generator of Pairwise-GAN is based on U-Net~\cite{ronneberger2015u} which is also used in the generator of Pix2Pix and CycleGAN~\cite{zhu2017unpaired,isola2017image}. Unlike usual conditional adversarial networks and ~\cite{huang2017beyond,liu2016coupled,anoosheh2018combogan}, Pairwise-GAN has two independent auto-encoders and weight sharing of the first two-layers of the decoder (Fig.~\ref{fig:pw_gan}). The layered architecture of U-Nets contains eight blocks in the encoder, eight blocks in the decoder, and skip connections between the encoder and decoder. Specifically, the detail layers used in each block of encoder and decoder are the same as Pix2Pix, except for the normalization layer. In Pairwise-GAN, we use instance normalization instead of batch normalization to boost the performance of generator~\cite{ioffe2015batch,ulyanov2016instance,wu2018group}.

When training, Pairwise-GAN requires two side-pose images at the same angles from the same person but in a different direction as inputs ($X_{left}$, $X_{right}$). Left generator $G_{1}$ takes the side-pose image from the left direction, while another side-pose image will be passed into generator $G_{2}$. After different encoders extract facial features, two frontal facial images are generated through two decoders. As these two generators are independent, only one side-pose image is required to be input for prediction. This approach addresses affordability and the difficulties in pixel-to-pixel transformation inherent in the complexities of human faces. 

\subsubsection{Discriminator}

\begin{figure}[t]
      \centering
      \includegraphics[height=4cm]{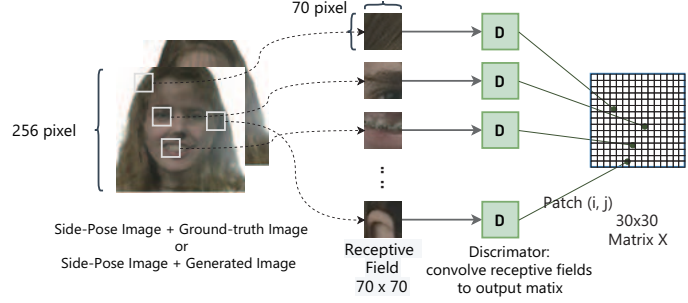}
      \caption{Intuitive view of the discriminator in Pairwise-GAN (Patch-based discriminator of GANs~\cite{isola2017image})}
      \label{fig:partchgan}
\end{figure}

 A standard discriminator in a GAN maps the input image into a scalar value which classifies whether it is real or fake. The output scalar value is a weighted sum of the whole data field, which cannot reflect the characteristics of the local feature. To address this lack of precision, we used the patch-based discriminator of GANs (PatchGAN) as the discriminator for Pairwise-GAN as proposed in Pix2Pix~\cite{isola2017image}. PatchGAN maps the input image (256 by 256 pixel) to an N-by-N matrix of outputs $X$, where each $X_{i,\  j}$ represents whether the patch $(i,\ j)$ is real or fake. Additionally, the patch $(i,\ j)$ is the convolution result of one receptive field which the discriminator is sensitive to, for example, the ear, tooth, and eyes if the input image is a human face (Fig.~\ref{fig:partchgan}). 

The operation described above is mathematically equivalent to cropping the input data into multiple overlapping patches, respectively discriminating the difference by the classifier (discriminator), and averaging the results. The result outlined in~\cite{isola2017image,zhu2017unpaired}, indicated that PatchGAN extracted the local characterization of the images. This is conducive to generating the images in a high resolution. 

From characteristics of patches, the size of the receptive field can be adjusted in different tasks. Specifically, a broader scope of receptive field focuses on the relationship between objects in a large area. However, it also consumes more expensive computation power. From the experimental results offered by Pix2Pix~\cite{isola2017image}, the receptive field with 16-by-16 size already achieves a sharp output. In contrast, results from a smaller receptive field (1-by-1) lose spatial statistics, and a larger receptive field (70-by-70) can force results with more colorfulness. As a result, under our computation power, we set the size of the receptive field as 70-by-70, which results in the output matrix being 30-by-30.

\subsubsection{Pair Loss}

To reduce the difference of generated images between the left generator and right generator, we examined the weight sharing between two encoders and added the pair loss into the loss function. This loss was expected to penalize the network parameters if there exists a major difference between generated images by the left generator and right generator. 
\begin{equation}
  Pair \ Loss = \frac{1}{W \times H} \sum_{i=1}^{W} \sum_{j=1}^{H} |\ YL_{i,\ j}  - YR_{i,\ j}\ | 
\end{equation}
where $YL$ and $YR$ refer to the image generated by the left generator and right generator respectively; $W$, $H$ is the width and height of the images; $i$, $j$ is the $x$ coordinate and $y$ coordinate of that pixel.

\section{Results and Evaluation}
\label{cha:result}

To measure the results quantitatively, we employed the facial similarity, structural similarity (SSIM), Frechet Inception Distance (FID), and Peak Signal-to-Noise Ratio (PSNR) by comparing the frontal facial synthesis and ground truths. Notably, The result of facial similarity (confidence) is provided by a commercial face comparison API (Face ++) based on face recognition. 

\subsection{Loss Function Analysis}

In this experiment, a Pix2Pix with various loss functions was trained with 125 epochs, with one epoch taking 2,090 pairs of images. Additionally, both generator and discriminator were optimized with Adam and a learning rate of 0.0002, following  ~\cite{ronneberger2015u}. 

Based on the default ratio of adversarial loss and L1 loss in Pix2Pix (1:120), we first adjusted the proportion of the L1 loss. With a low penalty coming from L1 loss, Pix2Pix tends to generate the frontal facial less similar to the original domain. If the L1 loss is removed from the loss penalty, the average similarity drops down 40$\%$ compared to the default one (Table~\ref{tab:loss_compare}). However, although a large L1 penalty results in a highly similar image, the blurring issue of synthetic frontal images becomes more serious. For instance, we found that the shape of the nose is difficult to determine in the generated image by 120 weights of L1 loss, compared to the adversarial loss one. 

After applying the GDL (gradient difference loss) into Pix2Pix and keeping the original ratio, the average similarity decreased. To further analyze the influence of the GDL penalty to model training, the proportion of L1 loss is decreased below the adversarial loss; the ratio between L1 and GDL remains untouched. We noticed that the average similarity grows to 35. However, the obscured side of the input image is blurred in the synthesis, which we considered a drawback of the GDL penalty.

The behavior of symmetry loss resulted in a negative influence on the performance of Pix2Pix as the average similarity decreases into 31. It was hard to converge during the training progress. As a result, we abandoned the this loss in later experiments. 

We explored identity loss which aims to regularize the generator. Compared to the L1 penalty and GDL penalty, identity loss leads to a significant improvement in the frontal facial of Pix2Pix, which the synthetic frontal face is most natural and sharp compared to others, notably, hairstyle and glasses were recovered in high quality. As a result, identity penalty helped to preserve facial identity information, however, adding identity loss to GAN learning doubled the training time required in each epoch to calculate the ground identity image through the generator. 

\begin{table*}[t]
    \centering
    \caption{Compare average similarity between ground truth and generated image on different loss configurations among 40 test images. Higher similarity is better}  
    \label{tab:loss_compare}
\begin{tabular}{cccccc|c}
\hline
 Hyper-parameters        & Adv        & L1           & GDL        & Sym        & Id         & Avgerage Similarity         \\ \hline
Pix2Pix                   & 1          & 0            & 0          & 0          & 0          & 25.68          \\
Pix2Pix                   & 1          & 40           & 0          & 0          & 0          & 33.11          \\
Pix2Pix                   & 1          & 120          & 20         & 0          & 0          & 31.17          \\
Pix2Pix                   & 20         & 3            & 0.5        & 0          & 0          & 34.89          \\
Pix2Pix                   & 20         & 3            & 0.1        & 0.05       & 0          & 31.07          \\
Pix2Pix                   & 20         & 3            & 0.1        & 0          & 5          & \textbf{37.92} \\ \hline
\textit{Pix2Pix Default}  & \textit{1} & \textit{120} & \textit{0} & \textit{0} & \textit{0} & \textit{35.2}  \\ \hline
\textit{CycleGAN Default} & \textit{1} & \textit{0}   & \textit{0} & \textit{0} & \textit{5} & \textit{38.7}  \\ \hline
\end{tabular}
\end{table*}

\subsection{Pairwise-GAN Analysis}

We trained each of the experiments of Pairwise-GAN for 250 epochs, with one epoch taking 1,045 images and used the same data distribution as previous experiments. We set the batch size to 1 and the learning rate to 0.0002. Additionally, Adam was selected as the optimizer for both generator and discriminator.  

We expected the network architecture with weight sharing gains to achieve a better performance than using the loss penalty as the former is closer to a coercive specification. However, the results indicated another option which the pair loss helps to improve the average performance, and weight sharing contributes to refresh the peak performance (Table~\ref{tab:Pairwise-GAN_results}). If both weight sharing and pair loss are employed in the Pairwise-GAN, it decreased both peak and average performance, which results from the network is hard to converge. 

\begin{table*}[t]
    \centering
    \caption{Results of different configurations on Pairwise-GAN (ablation study) and comparsions on Pairwise-GAN with Pix2Pix, CycleGAN, and CR-GAN}  
    \label{tab:Pairwise-GAN_results}
\begin{tabular}{cccccc|ccc|ccc}
    \hline
    \multicolumn{6}{c|}{Hyper-parameters}                                               & \multicolumn{3}{c|}{Similarity Measure}     & \multicolumn{3}{c}{Other Measure}                                    \\ \hline
    Adv         & L1         & GDL        & Id         & Pair        & Weight Share     & Avg            & Max            & Min                     & SSIM      & FID      & PSNR     \\ \hline
    10          & 0          & 0          & 5          & 10          & Disable          & 44.23          & 79.14          & 22.02                   & 0.576                   & 102.141          & 13.968                   \\ \hline
    \textit{10} & \textit{3} & \textit{0} & \textit{5} & \textit{10} & \textit{Disable} & \textit{44.30} & \textit{74.22} & \textit{\textbf{23.56}} & \textit{\textbf{0.600}} & \textit{91.517}  & \textit{\textbf{14.517}} \\ \hline
    10          & 3          & 0          & 5          & 0           & Enable           & 42.34          & \textbf{81.83} & 17.16                   & 0.550                   & \textbf{90.531}  & 14.199                   \\ \hline
    10          & 0          & 0          & 10-5       & 10-2        & Enable           & 42.79          & 70.61          & 16.43                   & 0.515                   & 93.433           & 13.987                   \\ \hline
    10          & 0          & 0          & 0          & 10-2        & Enable           & \textbf{45.16} & 71.95          & 20.07                   & 0.505                   & 119.211          & 13.209                   \\ \hline
    \multicolumn{6}{c|}{\textit{CR-GAN Default}~\cite{tian2018cr}}                                      & \textit{30.08}  & \textit{47.16} & \textit{12.9}          & \textit{0.362}          & \textit{171.28}  & \textit{10.916}          \\ \hline
    \multicolumn{6}{c|}{\textit{Pix2Pix Default}~\cite{isola2017image}}                                       & \textit{35.2}  & \textit{62.12} & \textit{0}              & \textit{0.502}          & \textit{118.791} & \textit{13.775}          \\ \hline
    \multicolumn{6}{c|}{\textit{CycleGAN Default}~\cite{zhu2017unpaired}}                                      & \textit{38.9}  & \textit{70.22} & \textit{14.64}          & \textit{0.493}          & \textit{99.777}  & \textit{13.274}          \\ \hline
    \multicolumn{6}{c|}{\textit{Groud Truth}}                                           & \textit{100}   & \textit{100}   & \textit{100}            & \textit{1}              & \textit{0.0}     & \textit{+Inf}            \\ \hline
\end{tabular}
\end{table*}

\subsubsection{Comparing with State-of-the-Art}

To further analyze the performance of Pairwise-GAN, we included comparisons with current state-of-the-art models. FT-GAN, the current state-of-the-art model in frontal facial generation used CMU Multi-PIE database~\cite{multi-pie}, and did not release the codes~\cite{zhuang2019ft}. Therefore, we compared our model with CycleGAN since FT-GAN is built on CycleGAN and within 4 $\%$ better than it. Additionally, we also the CR-GAN in the comparison.

Table~\ref{tab:Pairwise-GAN_results} shows Pairwise-GAN reaches 44.3 average similarity and 74.22 maximum similarity between synthesis and ground truth. In other words, the performance of Pairwise-GAN is around 5.4$\%$ better than the CycleGAN and 9.1$\%$ than the Pix2Pix, measured by the average similarity of synthesis and ground truth among 40 test images. It also obtains the best evaluation results from SSIM, FID, and PSNR. From the synthetic frontal images, we note that CycleGAN only generated the main facial features, and stacks onto the original side-pose image. On Pairwise-GAN, the field recovered is broader as it focuses on the facial features as well as other elements, such as hairstyle and neck. Apart from that, Pairwise-GAN also requires less computation power in both training and prediction as it only consumes 224 seconds compared to CycleGAN requiring 315 seconds. The prediction can achieve 34 fps by using GPU, RTX 2060S.

\begin{figure}[t]
      \centering
      \includegraphics[height=8cm]{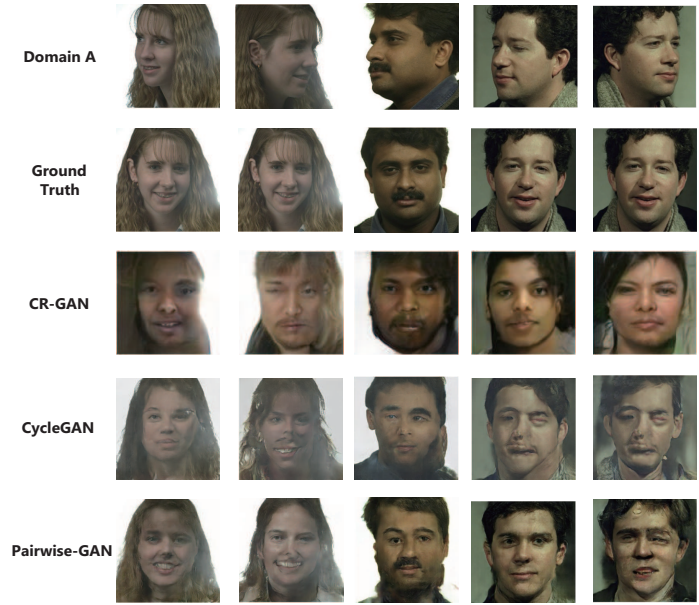}
      \caption{Qualitative results of Color FERET Database. Comparison with CR-GAN, CycleGAN and Ground Truth}
      \label{fig:Pairwise-GAN_samples}
\end{figure}

\section{Conclusion and Future Work}

In this paper, we extend current work on synthesizing front face images based on side-pose facial images. Initially, we focused on the performance of Pix2Pix, which tests and analyses five different loss functions, including adversarial loss, L1 loss, gradient difference loss, symmetry loss, and identity loss. Since the improvement based on different loss penalty was minor, we continued our work to propose Pairwise-GAN. Through the analysis of experimental results on different loss functions, we concluded that L1 loss, GDL loss, and identity loss help alleviate the common issue existing in CGAN, where the output face is related to the input face. The ratio of these five-loss, achieving the highest score in our experiment is 20:3:0.1:0:5 (adversarial:L1:GDL:symmetry:identity). 

In our second experiment, we proved that Pairwise-GAN generated better frontal face results than the CycleGAN in Color FERET database by using less computation resources; specifically, it improves around 5.4$\%$ on average compared to CycleGAN in generated quality. We also explored various configurations of Pairwise-GAN, which demonstrated that either, but not both, pair loss (soft penalty) or weight sharing (coercive specification) positively contribute to improvements. Further research on the network architecture is required to continue minimizing the difference of generated images by the two generators of Pairwise-GAN.

\section*{Acknowledgements}
We thank Dawn Olley and Alasdair Tran for their invaluable editing advice.

% \bibliography{mybibliography}%

\end{document}